\documentclass[manuscript,screen]{acmart}
\usepackage{caption}
\usepackage{subcaption}
\AtBeginDocument{%
  \providecommand\BibTeX{{%
    \normalfont B\kern-0.5em{\scshape i\kern-0.25em b}\kern-0.8em\TeX}}}

\setcopyright{acmcopyright}
\copyrightyear{2018}
\acmYear{2018}
\acmDOI{10.1145/1122445.1122456}

\acmConference[THRI]{Transactions on Human-Robot Interaction}
\acmBooktitle{Transactions on Human-Robot Interaction}

\begin{document}

\title[An AR-HMD Interface for Human Robot Collaboration]{Design and Evaluation of an Augmented Reality Head-Mounted Display Interface for Human Robot Teams Collaborating in Physically Shared Manufacturing Tasks}
\titlenote{Initial partial results published in \cite{chan2020AR}}

\author{Wesley P. Chan}
\affiliation{%
  \institution{Monash University}
  \country{Australia}
}

\author{Geoffrey Hanks}
\affiliation{%
  \institution{University of British Columbia}
  \country{Canada}}

\author{Maram Sakr}
\affiliation{%
  \institution{University of British Columbia}
  \country{Canada}}

\author{Haomiao Zhang}
\affiliation{%
  \institution{University of British Columbia}
  \country{Canada}}

\author{Tiger Zuo}
\affiliation{%
  \institution{University of British Columbia}
  \country{Canada}}

\author{H.F. Machiel Van der Loos}
\affiliation{%
  \institution{University of British Columbia}
  \country{Canada}}
  
\author{Elizabeth Croft}
\affiliation{%
  \institution{Monash University}
  \country{Australia}
}

\renewcommand{\shortauthors}{Chan et al.}

\begin{abstract}
We provide an experimental evaluation of a wearable augmented reality (AR) system we have developed for human-robot teams working on tasks requiring collaboration in shared physical workspace. Recent advances in AR technology have facilitated the development of more intuitive user interfaces for many human-robot interaction applications. While it has been anticipated that AR can provided a more intuitive interface to robot assistants helping human workers in various manufacturing scenarios, existing studies in robotics have been largely limited to teleoperation and programming. Industry 5.0 envisions cooperation between human and robot working in teams. Indeed, there exist many industrial task that can benefit from human-robot collaboration. A prime example is high-value composite manufacturing. Working with our industry partner towards this example application, we evaluated our AR interface design for shared physical workspace collaboration in human-robot teams. We conducted a multi-dimensional analysis of our interface using establish metrics. Results from our user study (n=26) show that subjectively, the AR interface feels more novel and a standard joystick interface feels more dependable to users. However, the AR interface was found to reduce physical demand and task completion time, while increasing robot utilization. Furthermore, user's freedom of choice to collaborate with the robot may also affect the perceived usability of the system. 

\end{abstract}




\maketitle

\section{INTRODUCTION}
        \label{sec:introduction}
        The introduction of robotic devices has yielded significant advancements and productivity gains in manufacturing. However, although many manufacturing processes are now fully automated, there remains a significant number of processes that are still performed manually. Such tasks are inherently difficult to automate due to their complex and high-variability nature. These tasks typically required the dexterity, expertise, and cognitive capabilities of humans; capabilities that are not yet achievable by current industrial robots. These processes often become the rate limiting step in the entire manufacturing process. To improve the productivity of such processes, the introduction of human-robot teams where robotic assistants collaborate with human workers has attracted great interest. Still, attempts in deploying robotic assistants within human populated productions lines to form human-robot teams have yielded mixed success. One challenge is the lack of an intuitive interface for communicating with the robot. Current methods for programming and communicating with industrial robots through teach pendants or computer consoles \cite{Daniel2013} are unintuitive, complex, inflexible, and do not provide a suitable modality for online interaction and collaboration. Such interfaces discourage interaction with the robotic partner, distract the workers from the task, and compromise the team's performance. Productivity and safety concerns also arise when the worker’s attention and focus are taken away from the task by diverting attention to the user interface. To enable effective human-robot teams in manufacturing, more intuitive and task-focused interfaces for communicating with robot partners are required.

\begin{figure}[t]
\centering
\includegraphics[width=.35\textwidth]{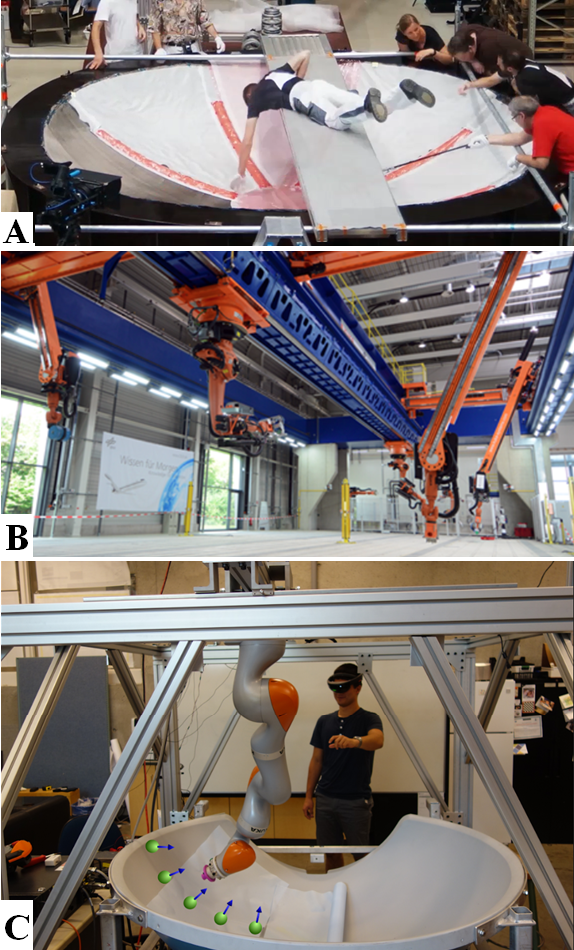}
\vspace*{-3mm}
\caption{A - Pleating procedure in CFRP manufacturing. B - DLR's factory in Augsburg with large ceiling-hanging robots. (Images from DLR.) C - Robot setup in the lab. User collaborating with the robot teammate using our AR system. (Published in \cite{chan2020AR}.)}
\vspace*{-7mm}
\label{fig:overview}
\end{figure}

A particular class of manufacturing procedures where the introduction of human-robot teams can benefit worker ergonomics, task quality and productivity is large-scale, labour-intensive tasks. Such tasks have high physical demands as workers are required to move around and operate on large workpieces. A representative example of such procedures is the pleating procedure for carbon-fibre-reinforced-polymer (CFRP) fabrication for aerospace applications. 
We have been investigating solutions for enabling human-robot teams with our collaborator, the German Aerospace Center, DLR, for CFRP fabrication at their facility in Augsburg.
CFRP manufacturing requires workers to first apply layers of fabric onto a mould, followed by a membrane covering prior to injecting resin to create the part. To ensure a smooth result, workers need to form pleats on the membrane covering, and move them to specific locations of the mould to remove all wrinkles on the membrane. This pleating process requires the human worker’s dexterity and expertise in knowing which direction to move the pleats depending on where the wrinkles form. 
Since aerospace workpieces can be very large, the pleating task is physically demanding, as it requires workers to climb over scaffolds to reach every part of the workpiece (Fig. \ref{fig:overview}A). Large-scale robots at DLR's factory (Fig. \ref{fig:overview}B) are proposed to serve as assistants to the human workers, with the aim of reducing worker physical demand and improving task repeatability and efficiency. Due to the size, weight, and power of these robots, safety becomes a significant concern, and other programming and interaction methods, e.g., kinesthetic teaching \cite{CakmakKinTeach}, are infeasible. To be able to realized the benefits such robotic team members can potentially offer, we need an alternative interface that can facilitate safe and intuitive communication and interaction with human workers that are non-robotics experts.

Recently advancements in augmented reality (AR) technology have made AR an attractive candidate method for enabling intuitive human-robot interaction. Through the use of AR, we can create co-located user interface components by rendering virtual objects over in the physical workspace. Such an interface would permit the user to maintain focus on the task while commanding or interacting with the robot. Furthermore, commercial AR devices are becoming increasingly available \cite{MagicLeap, kress201711, Moverio}, with most supporting natural interaction methods using gestures, speech, and/or gaze. These together allow creation of visually rich, intuitive user interfaces.
In this paper, we investigate the use of AR and its benefits for human-robot teams collaborating in shared space in manufacturing scenarios. We present a user study on the use of an intuitive AR interface for instructing and collaborating with a robotic assistant partner in an experiment task simulating CFRP fabrication.

\subsection{Paper Evolution}
Early, partial results of a smaller user study on our AR system were published in \cite{chan2020AR}. This paper provides the following key new contributions.

\begin{enumerate}
    \item \textbf{Additional experimentation.} We conducted additional experimentation with 16 additional participants, increasing our sample size from n=10 \cite{chan2020AR} to a total of n=26. The results from all 26 participants are reported in this paper in Section \ref{sec:results}.
    \item \textbf{More detailed analysis.} In addition to analyzing task completion time, robot utilization, and NASA Task Load Index (NASA-TLX) responses as in \cite{chan2020AR}, we provide further analysis of additional measures including the System Usability Scale (SUS) and the User Experience Questionnaire (UEQ), and compare with established benchmarks, providing a more in-depth and multi-dimensional evaluation of our AR interface for human-robot collaboration. Results of the analysis of these additional measures are presented in Section \ref{sec:results_subjective_measures}, and discussed in Sections \ref{sec:discussion_sus_ueq} and \ref{sec:discussion_novelty}.
    \item \textbf{Corroborating evidence.} Results of the analysis from our current larger study (n=26) have confirmed and strengthen our previous findings reported in \cite{chan2020AR} with a smaller sample size (n=10). Our initial finding and conclusions are largely confirmed, with higher statistical significance shown in current results (Section \ref{sec:results}). Results from the additional measures further support the benefits of our AR system for humans-robot collaboration from a system usability and user experience standpoint (presented in Section \ref{sec:results_subjective_measures} and discussed in Section \ref{sec:discussion_sus_ueq}).
    \item \textbf{Recommendations for AR-based human-robot teams.} Based on the findings of our larger user study, this paper also provides recommendations for implementing AR-based human-robot teams in Section \ref{sec:discussion_recommendations}. 
\end{enumerate}

\section{RELATED WORK}
        \label{sec:related_works}
        Recently, there have been many applications of AR to a range of tasks including training~\cite{webel2013augmented}, 
assembly~\cite{wang2016comprehensive}, 
repair~\cite{henderson2011exploring},
and 
maintenance~\cite{engelke2015content}, and the results these studies have motivated a growing range of applications. In this section, we provide a review of the literature related to the use of AR for robotics.

\subsection{AR Robot Surrogates}
One of the earliest applications of AR in robotics was to enable of interaction and control of a real robot, through a virtual proxy of the robot visualised in AR. Chong et al. \cite{chong2009} created a robot guidance system enabling users to control a real robot through interaction with a virtual copy. Another system was created by
Fang et al. \cite{fang2014novel}. They implemented their system using a monitor display, and users could program a real robot by moving a virtual robot rendered onto the monitor. In their study, it was found that the use of a monitor display reduced depth perception, and distracted the user's attention from the task. 
An AR system for controlling drones was created by Walker et al. \cite{walker2019}. Their system enabled robot control via virtual robot surrogates visualised in AR. They reported improved completion time and subjectively reduced stress levels with the use of their AR system.

\subsection{AR as In-Situ Displays}
Another group of literature focuses on the use of AR for displaying robot information to the user. Andersen et al. constructed a system for visualizing task and robot information using projection AR for a vehicle door assembling task \cite{andersen2016projecting}.
Ro et al. equipped their robot with the capability of projecting arrows onto the floor for guiding users to their destination \cite{ro2019}, while Reardon et al. created a similar guidance system using a head mounted AR device \cite{reardon2018}. Lim et al.
presented a system for mini-car driving, combining the use of both a smart phone and a projector to provide an enhanced user-immersion experience \cite{Lim2015}.
Kemmoku and Komuro considered applications requiring a large effective area of display, and they built an AR display system using a projector mounted onto the user's head to achieve this \cite{Kemmoku2016}. Projection-based AR systems, however, suffer from occlusions by objects in the environment or by the user themselves, rendering them unsuitable for  human-robot collaboration scenarios.
To avoid occlusion issues, Hanson et al. built an AR system using a head-mounted device \cite{hanson2017augmented}. Their system targeted an assembly tasks and was used to display assembling instructions to the user. They reported that their AR interfaces increased task efficiency and accuracy; however, there were no robots involved in this study. Rosen et al. created a system that uses AR displays for bidirectional human-robot communication \cite{rosen2020}. Their application focused on communication and did not involve collaboration in shared space. Walker et al. created a system using a head-mounted display to show a drone's future path/waypoints. They compared different types of visualization in a task involving human and robot sharing the same workspace, but not the same goal \cite{walker2018}.
In summary, studies in this category focus on the use of AR as a communication tool for providing information to the user. However, they do not involve shared-space interaction or collaboration between human and robot partners working together toward a shared goal.

\subsection{AR for Teleoperation and Programming}
The majority of literature on AR systems for human-robot interaction have used AR as a robot teleoperation or programming interface. Rosen et al. presented an AR system that allows users to set robot arm goal poses, and provides robot motion preview \cite{rosen2019}.
Ni et al. presented a system that uses AR and a haptic device for a welding task application \cite{ni2017haptic}. 
Their system enabled users to specify a remote robot's welding trajectory through a monitor displaying the remote scene and an overlaid virtual robot. Gregory et al. presented system that utilizes a head mounted AR display along with a gesture glove that allows the user to command a mobile robot \cite{gregory2019}. The robot was equipped with a certain level of autonomy for reconnaissance missions. 
Stadler et al. analyzed the workload on industrial robot programmers as they controlled a robot with a tablet-based AR interface \cite{stadler}. They reported a decrease in mental demand, but increase in completion time. Their study involved a miniature robot (Sphero). Quintero et al. also created an AR system that allows user to program, preview, and edit robot trajectories \cite{Quintero2018}. They tested their system with a table-top robotic arm, and instead they found that their AR system resulted in increased mental demand, but reduced robot teaching time. 
Similarly, Ong et al. also found that an AR-based interface allows users to more quickly and intuitively program a table-top robot for welding and pick and place tasks. \cite{ONG2020101820}, while Frank et al. found that using a tablet-based AR system to instruct a table-top robot for a pick and place task yields more efficient performances \cite{Frank2016}. 
The variations in results reported from these studies  suggest that robot size and task type can affect the performance of AR interfaces.

Although the benefits of AR for improving task efficiency in various robotic applications have been demonstrated, most studies have been limited to robot teleoperation, control, or programming tasks, where users are restricted to interaction with AR objects in virtual space, while the robots operate separately in the physical workspace. Furthermore, existing studies have been largely limited to table-top-sized robots and low-physical-demand tasks. There have not yet been studies focusing on human-robot teaming applications requiring both human and robot partner to collaborate in the same shared workspace, working on the same physical workpiece simultaneously. Further, there have not been studies focusing on larger-scale robots, or high-physical-demand tasks. Industry 5.0 focuses on human-machine cooperation, and there are numerous industrial applications where human-robot collaboration in a shared workspace is of particular interest \cite{Gleeson2013,Wilcox2012}. Hence, towards enabling effective human-robot teams in CFRP manufacturing, we conducted a user study to investigate and evaluate the performance of an AR interface we developed for human-robot teams, working collaboratively on a high-physical-demand shared task with a shared workpiece and shared workspace.

\section{OBJECTIVE}
        \label{sec:Objective}
        
Our objective is to investigate the potential for AR technology to provide an intuitive and effective interface for facilitating efficient human-robot teams. In particular, we are interested in the context of human-robot collaboration in large-scale, labour-intensive manufacturing tasks, such as CFRP manufacturing (Fig. \ref{fig:overview}A).
In such tasks, human and robot partners will need to work together on the same workpiece in the same physical workspace. Our guiding research questions are: 
\begin{itemize}
  \item Can the use of an AR-based interface for human-robot collaboration increase overall task efficiency?
  \item What are the effects of an AR-based interface on the perceived task load?
  \item Can the use of an AR-based interface encourage human-robot collaboration and promote robot utilization?
\end{itemize}
To seek answers for these questions, we conducted a user study to compare the use of an AR interface we built, with a standard joystick interface, for an experimental task simulating collaborative CFRP manufacturing.

\section{SYSTEM}
        \label{sec:system}
        \subsection{Robot Platform}
Our robot platform is shown in Figure \ref{fig:overview}C. We constructed a robotic test bench with a two-axes movable platform. The test bench measures approximately 1.8 m x 1.7 m x 1.9 m. A KUKA IIWA LBR14 robot arm is mounted on the movable platform, and below the robot arm is the workpiece. The KUKA IIWA has impedance control capabilities, providing a suitable platform for performing tasks requiring physical contact with the environment and safe interaction with humans. The two-axis movable platform serves to move and position the robot arm around the workspace, extending the robot arm's reachable range, such that it can reach all parts of the workpiece. At our partner DLR's factory, the KUKA IIWA will be mounted on their large ceiling-hanging robots (Figure \ref{fig:overview}B), which will serve to move and position the KUKA IIWA arm. In our user study, we attached a 3D printed spring-loaded mechanism onto the wrist of the KUKA IIWA arm for mounting the tool (a marker pen). The spring-loaded mechanism serves to provide passive compliance for safe contact with the workpiece.

\subsection{AR Human Robot Collaboration Interface}
\begin{figure*}
\vspace*{1mm}
\centering
% \subfigure
{   \includegraphics[trim = 0mm 0mm 0mm 0mm, clip=true,width=0.8\textwidth] 
{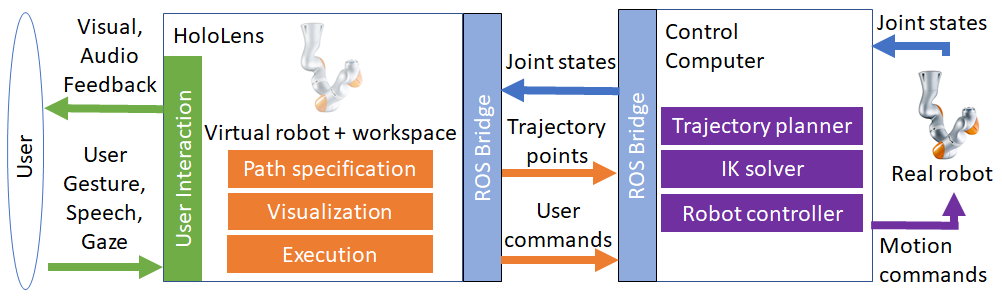} 
}
\vspace*{-5mm}
\caption{System block diagram - The user interacts with the HoloLens through gesture, speech, and gaze, while visual and audio feedback is provided. Our AR interface provides path specification, visualization, and execution functionalities. The HoloLens communicates with control computer using ROS Bridge \cite{rosbridge}. The control computer commands robot using the motion planner library MoveIt \cite{moveit}. The real robot continuously sends current joint states to control computer, which then sends it to HoloLens for visual display as feedback to user. (Published in \cite{chan2020AR}.)
}
\vspace*{-7mm}
\label{fig:system_block}
\end{figure*}

Figure \ref{fig:system_block} shows the block diagram of our AR human-robot collaboration system. Our system consist of three main components. A Microsoft HoloLens AR head-mounted display \cite{kress201711}, a robot control computer, and communication library ROS Bridge \cite{rosbridge}. The HoloLens is used to create an immersive, intuitive user interface. Unity, along with the programming library ROS\# \cite{ros-sharp} is used to develop the AR interface. The Hololens renders 3D AR displays and user interface elements, co-located with the real robot and workspace, for user interaction. It permits user inputs including speech, arm gesture, and gaze. During operation, our AR systems renders a geometrically accurate robot and workpiece model to show to the user. Different models can be loaded depending on the current robot partner and workpiece. A fiducial marker (ARTag) placed at a known position relative to the robot and the workpiece is used for calibrating the positions of the virtual models with the real counterparts. The rendered virtual robot and workpiece models provide tasks context and visual feedback of the calibration result to the user. Our system allows the user to specify the robot path, visualise the motion, and execute robot trajectories. The robot control computer processes user commands, completes the necessary trajectory planning and inverse kinematics, and sends low level commands to the real robot for motion execution. The Robot Operating System (ROS) is used to implement these functionalities. The communication library ROS Bridge is used to transmit data between the Hololens and the robot control computer through a wireless connection, and and manages the necessary translation between different data formats.

\begin{figure*}
\centering
% \subfigure
\vspace*{2mm}
{   \includegraphics[trim = 0mm 0mm 0mm 0mm, clip=true,width=0.9\textwidth] 
{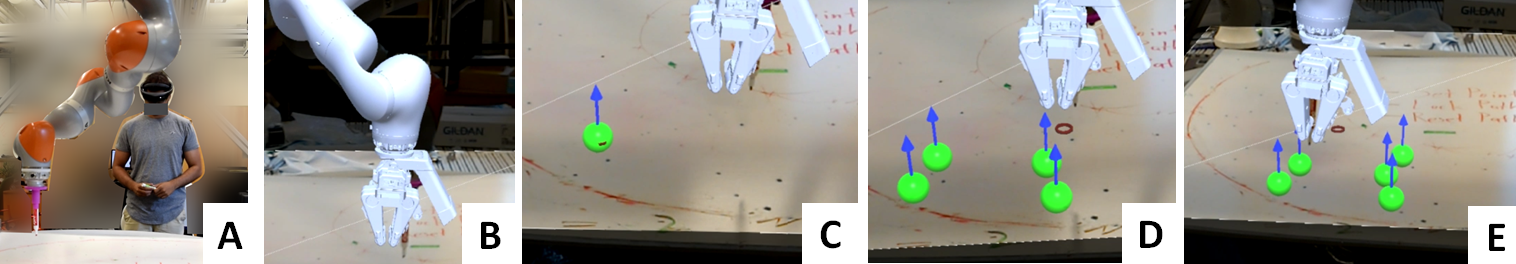} 
}
\vspace*{-3mm}
\caption{Use of our AR system for programming robot trajectories. A - The user collaborates with robot partner through our AR interface. B - A geometrically accurate model of the robot is rendered over the real robot and displayed to the user. C - The user sets a way point through gaze and speech. D - Multiple way points are set to define a path. E - The user commands the robot to execute the set path with a speech command. (Published in \cite{chan2020AR}.)
}
\vspace*{-7mm}
\label{fig:AR_system}
\end{figure*}

Figure \ref{fig:AR_system} shows the workflow for using our AR system to specify and execute the robot's motion. Unlike standard teach pendents that distract user attention away from the robot, our system facilitates human-robot interaction through interface components rendered in AR, co-located in the same physical space as the robot and the workspace (\ref{fig:AR_system}A). A virtual model of the robot and the workpiece is rendered over the real robot and workpiece to provide a visual indication of the positional calibration result and task context (Figure \ref{fig:AR_system}B). At anytime, if the user notices any positional mismatch (due to drift in HoloLen's spatial tracking), they can look at the ARTag marker to recalibrate the system. The system infers the gaze location of the user by tracking a ray from the person's head orientation. A ring marker is rendered on the workpiece at the intersecting point with this ray to indicate the gaze location of the user. The user can set trajectory way points at the marker location by using the speech command "set point". A green sphere with a blue arrow indicating the surface normal is then rendered to indicate the set point location (\ref{fig:AR_system}C). By repeating this process, the user can set successive trajectory way points on the workpiece (\ref{fig:AR_system}D). At anytime, the user can use the speech command "reset path" to clear any points that have been set. Once the user has finished setting all way points of the trajectory, they can then use the speech command "execute" to instruct the robot to execute the trajectory they have set (\ref{fig:AR_system}E). During trajectory execution, the virtual model reflects the motion of the real robot. This provides a confirmation of the communication connectivity between the HoloLens and the control computer to the user.

\subsection{Joystick Trajectory Programming Interface}
We also created a joystick interface for controlling and programming our robot, to serve as a proxy for the teach pendants used as the current industry standard inteface \cite{Daniel2013}. We used a standard PS3 controller to implement the joystick interface. Using this interface, the user controls the robot end effector's forward/backward, left/right motions by pushing the controller joystick up/down, left/right. A trajectory way point can be set at the current end effector location by pressing a set point button. A second button enables clearing of set way points. Pressing a third button commands the robot arm to execute the set trajectory. This robot programming modality provided by the joystick is analogous to the current programming methods supported by teach pendants.
        
\section{EXPERIMENT}
        \label{sec:experiment}
        \subsection{Participants}
We recruited participants for our user study through word of mouth, advertisements posted on the university campus, and social media. A total of twenty-six participants (nineteen male, seven female) participated in our user study. Prior to conducting the study, we obtained research ethics approval from the University of British Columbia Behavioural Research Ethics Board (application ID H10-00503). We obtained informed consent from each participant before commencing each experiment session.

\subsection{Experiment Task}
In our experiment, participants carried out a physical collaborative task, simulating CFRP manufacturing procedure\footnote{demo video, originally published along with \cite{chan2020AR}: https://youtu.be/roOKjVLS-Rc}. In a CFRP manufacturing task, the pleating procedure involves two main tasks. The first task is to start forming pleats around the edges of the mould, by gathering and neatly folding excess membrane material. The second task is to move and align the pleats along stringers that extend from the edges of the mould to the centre. The edge tasks requires more dexterity, while the centre tasks are more physically demanding. As the moulds we considered can reach 4 m in diameter and be concave, in order for the workers to reach the center parts of the mould, they need to setup scaffolds and climb over them in order to reach the center parts (Figure \ref{fig:overview}A). To alleviate the labour demand and risk associated with the task, in the envisioned human-robot collaboration scenario, it is intended that a human worker would perform the edge tasks, while the robot assistant would perform the center tasks. 

For our experimental task, to simulate the physically demanding nature of the task, we used a mould that is 1.6 m in diameter and 0.6 m in depth. The robot setup and the mould used in our experiment is shown in Figure \ref{fig:overview}C. This necessitated participants to move around the mould to perform all the edge task, and setup scaffolds to reach the center parts of the mould. As the pleating procedure is a complex tasks requiring expertise, for our experiment, we placed a flat whiteboard over the mould, and we asked participants to colour pre-defined lines on the whiteboard instead. Edges tasks were simulated with colouring of zig-zag lines, while center tasks were simulated with colouring of lines running from the edges to the center. To ensure safety, participants were asked to set up a scaffold as a stepping stool for reaching/performing the center tasks, instead of climbing over the scaffold or the mould. We asked participants to perform a total of 4 sets of pleating paths (edge path + center path) in a predefined order in each experiment. In a real CFRP manufacturing task, the worker must assemble and insert a vacuum tube under the membrane material in preparation for the next step following the pleating procedure. Participants were ask to assemble the actual tube as the last step in the experiment task as well.

\subsection{Experiment Conditions}
We wanted to test whether AR-enabled human-robot teams for physically shared tasks would provided benefits over the current manual method, and we wanted to test if our AR interface would provided benefits over existing alternative interfaces. Furthermore, we wanted to test whether our AR interface provides an effective interaction method that can encourage human-robot teaming and promote the utilization of the robotic assistant partner. Hence, we designed the following five conditions for our experiment:

\textbf{H: Human Only Condition.} The participant performs the entire task by themselves, without the use of the robot assistant. This condition serves as a baseline comparison and represents the current manual pleating process for CFRP fabrication.

\textbf{J1: Joystick Condition - Task Division Predefined.} The participant performs the experiment task collaborating with the robot assistant using the joystick interface. The joystick interface is analogous to current standard teach pendants. This allows us to compare our AR interface with an alternative standard interface. Considering the target application scenario as mentioned in the previous subsection, the human is designated to take on the easier-to-reach edge tasks (half of the task), and use the robot for the harder-to-reach, physically demanding center tasks (the other half of the task).

\textbf{AR1: AR Condition - Task Division Predefined.} The participant performs the experiment task collaborating with the robot assistant using our AR interface. Again, considering the target application scenario, participants are to complete half of the tasks, the edge tasks, while using the robot assistant to complete the other half of the tasks, the center tasks.

\textbf{J2: Joystick Condition - Task Division Unspecified.} The participant performs the experiment task collaborating with the robot assistant using the joystick interface. However, participants are given the freedom to choose which tasks to perform themselves, and which tasks to use the robot to perform. This freedom to decide when and how much to use the robot allows us to examine if the joystick interface affects robot utilization.

\textbf{AR2: AR Condition - Task Division Unspecified.} The participant performs the experiment task collaborating with the robot assistant using our AR interface. However, participants are given the freedom to choose which tasks to perform themselves, and which tasks to use the robot assistant's help to perform. This freedom to decide when and how much to use the robot allows us to examine if our AR interface affects robot utilization.

\subsection{Procedure}
At each experiment session, we first gave an overview of the pleating procedure and explained our experiment task to the participant. We explained to the participant how to use the joystick and the AR interfaces to control and program the robot. Then, we allowed the participant to try using the two interfaces. After the participant  familiarized themselves with interfaces, we then proceeded to the experiment trials. 

Each participant first performed the task in the H (Human Only) condition as a baseline. After that, the participant performed the task in the J1 and AR1 conditions. We counterbalanced the ordering of J1 and AR1 among participants to mitigate carryover effects. Finally, the participant performed the task in the J2 and AR2 conditions. We again counterbalanced the ordering of J2 and AR2 among participants to mitigate carryover effects. After performing the task in each condition, each participant was asked to complete a set of questionnaires (details in Section \ref{sec:analysis}) to evaluate the system. After experiencing all conditions, we also asked the participant to provide any additional comments they have, at the end of the experiment.          
\section{HYPOTHESES}
        \label{sec:hypotheses}
        We posit that the use of a robotic assistant and human-robot collaboration in physically shared manufacturing tasks can provide the benefits of reducing worker task load and increasing task efficiency. Furthermore, we believe that our AR interface will be able to provided a more intuitive and effective way of interacting and collaborating with a robot assistant, and hence, promote acceptance of the system and elicit higher robot utilization by the human workers. Based on these believes, we formulate the following hypotheses for our experiment:

\begin{itemize}
\item \textbf{H1}: Collaboration with a robot assistant, regardless of interface (J1, AR1, J2, AR2), reduces the physical demand on the worker, and shortens the task completion time, when compared with human only condition (H).

\item \textbf{H2}: Use of AR interface (AR1, AR2) shortens the task completion time, and reduces the task load on the worker, when compared to standard joystick interface (J1, J2).

%\item \textbf{H2}: The use of AR (AR1, AR2) compared to joystick (J1, J2) increases task efficiency and reduces completion time.

%\item \textbf{H3}: The use of AR (AR1, AR2) compared to joystick (J1, J2) reduces task load on the user.

\item \textbf{H3}: AR interface (AR1, AR2) improves system usability and user experience when compared to standard joystick interface (J1, J2).

\item \textbf{H4}: The AR interface (AR2) provides a better method for interacting with the robot, promoting human-robot collaboration, and increases robot utilization.
\end{itemize}

The first hypothesis \textbf{H1} examines the benefits of a \textit{robotic assistant} (regardless of interface) for manual manufacturing tasks. The second hypotheses \textbf{H2} examines the benefits of an AR interface on \textit{task efficiency} in terms of worker load and completion time. \textbf{H3} examines how an AR interface influences \textit{user perception} of the system, while \textbf{H4} examines the influences on \textit{user acceptance} of the system. (\textbf{H4} was not explicitly examined in \cite{chan2020AR} due to space limitations.)
        
\section{ANALYSIS}
        \label{sec:analysis}
        When evaluating new user interfaces, it is important to examine the system from both an objective and subjective point of view. Objective measures allow us to determine if the new interface brings measurable performance improvements to the process. Subjective measures, on the other hand, allow us to determine how the user perceives the use of the interface (such as perceived task load and usability). Subjective measures are important since how users perceive the interface relates to how likely they will accept the technology, which in turn determines if the technology will be continually utilized and further developed \cite{Klamer2010, Heerink2006, Moradi2018} We use the following objective and subjective measures to provide a multi-dimensional evaluation of our interface.

\subsection{Objective measures}

\textbf{Task Completion Time.} We measured the total time, $t$, required by the user to complete the experiment task in each condition. This provides a measurement of task efficiency. 

\textbf{Robot Utilization.} For the conditions where the participant was given freedom to choose how much to utilize the robot assistant (J2, AR2), We measured robot utilization by calculating the percentage of discrete paths (edge or center) executed by the robot among all eight paths to be executed by the human-robot team. This provides an indication of robot/system acceptance by the user. 

We compared completion time among conditions using ANOVA. Post hoc analysis was carried out using t-tests with Holm-Bonferroni method. To compared robot utilization, we used t-tests to determine if robot utilization in J2 and AR2 has either increased or decreased significantly relative to J1 and AR1, when robot utilization was pre-defined at 50\%. The alpha level was set to 0.05 for all analyses. 

\subsection{Subjective measures}
\textbf{NASA Task Load Index (NASA-TLX).} We used the NASA-TLX \cite{NASA-TLX} to measure the subjective task load on participants in each condition. The NASA-TLX is a questionnaire composed of six questions asking the participant to rate their experienced task load in six different aspects on a 21-point scale. 

\textbf{System Usability Scale (SUS).} To evaluate the usability of the system in each condition, we used the SUS \cite{SUS}. The SUS includes ten questions asking the participant to rate the system's usability in different aspects on a 10-point scale. An overall score is then calculated. The benefits of using the SUS is that it is a well established scale with benchmarks from studies in different types of user interfaces and applications, allowing us to obtain a general indication of system usability relative to other existing interfaces, not limited to AR or robot \cite{Sauro2012}. 

\textbf{User Experience Questionnaire (UEQ).} We also evaluated the user's experience on using the system in each condition via the UEQ \cite{UEQ}. The UEQ contains 26 questions on a 7-point scale, evaluating the participant's users experience on the systems from different aspects relating to six attributes including attractiveness, dependability, and novelty. Existing studies on the use of AR for robots do not often measure novelty. However, as wearable AR systems have not yet become widespread, it is likely that there will be novelty effects with AR-based interfaces. Hence, it is important to evaluate these related aspects as well. Like the SUS, the UEQ is a well-establish questionnaire with existing benchmarks, allowing comparison with other types of systems \cite{Schrepp2017}. 

Following existing studies \cite{stadler, Quintero2018, winter2010}, we performed ANOVA on the subjective measures to identify significant differences among conditions. Post hoc analysis was conducted using paired t-tests with Holm-Bonferroni method. The alpha level was set to 0.05 for all analyses.

\section{RESULTS}
        \label{sec:results}
        \subsection{Objective Measures}

\begin{table}[]
\vspace*{-3mm}
\caption{Completion time, ${t}$, measured in each condition.}
\vspace*{-3mm}
\label{table:time}
\centering
\begin{tabular}{ |c|c|c|c|c|c| } 
\hline
 & \textbf{H} & \textbf{J1} & \textbf{J2} & \textbf{AR1} & \textbf{AR2} \\
\hline
\textbf{$t$(s)} & $299\pm86$ & $244\pm62$ & $209\pm57$ & $181\pm48$ & $148\pm35$ \\ 
\hline
\end{tabular}
\vspace*{-1mm}
\end{table}

\begin{figure}[]

\centering
\includegraphics[width=.45\textwidth]{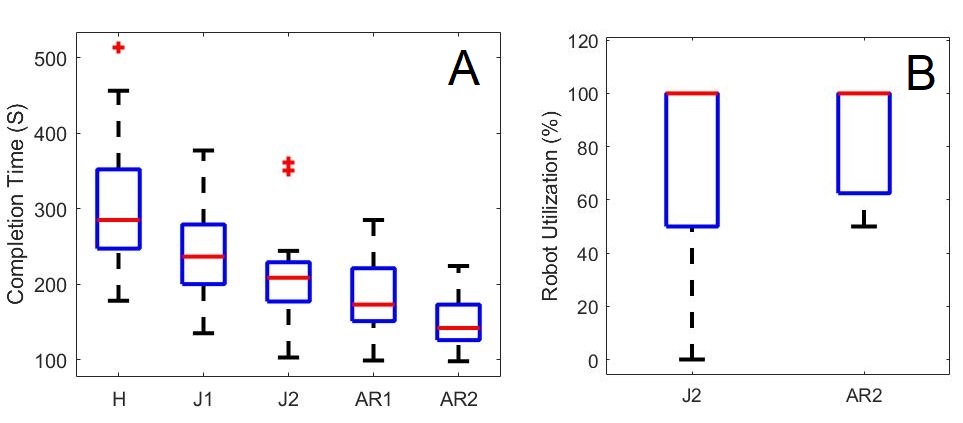}
\vspace*{-3mm}
\caption{A. Measured completion time, ${t}$. B. Measured robot Utilization, ${R}$.}
\vspace*{-3mm}
\label{fig:time_utilization}
\end{figure}

\textbf{Completion Time.}
Table \ref{table:time} reports the measured completion time, $t$, in each condition while Fig. \ref{fig:time_utilization}A shows the box plots. Overall, AR1 and AR2 were found to have the shortest completion times. ANOVA revealed that there were significant differences among the measured completion times ($F(4, 125)=24.6, p<0.001$). Pairwise t-test indicated that completion times are significantly shorter in J1, J2, AR1, and AR2 when compared to H ($t(25)=6.15,p<0.001$; $t(25)=7.00,p<0.001$; $t(25)=10.2,p<0.001$; $t(25)=9.74,p<0.001$). Furthermore, 
completion times in AR1, AR2 are also significantly shorter than that in J1, J2, respectively ($t(25)=8.10,p<0.001$; $t(25)=7.36,p<0.001$), and completion times in J2, AR2 significantly shorter than that in J1, AR1, respectively ($t(25)=5.14,p<0.001$; $t(25)=4.40,p<0.001$).

\textbf{Robot Utilization.}
Fig. \ref{fig:time_utilization}B reports the measured robot utilization, $R$, in J2 and AR2.
For J2, robot utilization ranged from 0\% to 100\%, and for 8 out of 26 participants, we measured a robot utilization of 50\% or less (with 4 having a robot utilization =50\% and 4 having a robot utilization <50\%). With the joystick interface, a number of participants decided to utilize the robot assistant less, or even not at all. For AR2, robot utilization ranged from 50\% to 100\%, and for 24 out of 26 participants, we measured a robot utilization of >50\%. This indicates that with the AR interface, all but two participant utilized the robot assistant more. A t-test revealed that, comparing with J1 and AR1, where robot utilization was set at 50\%, robot utilization overall increased in both J2 ($R=75.5\%\pm33.6$, $t(25)=11.4,p<0.001$) and AR2 ($R=84.6\%\pm19.5$,$t(25)=22.2,p<0.001$).

\subsection{Subjective Measures}
\label{sec:results_subjective_measures}

\begin{figure}[]
\centering
\vspace*{2mm}
\includegraphics[width=.75\textwidth]{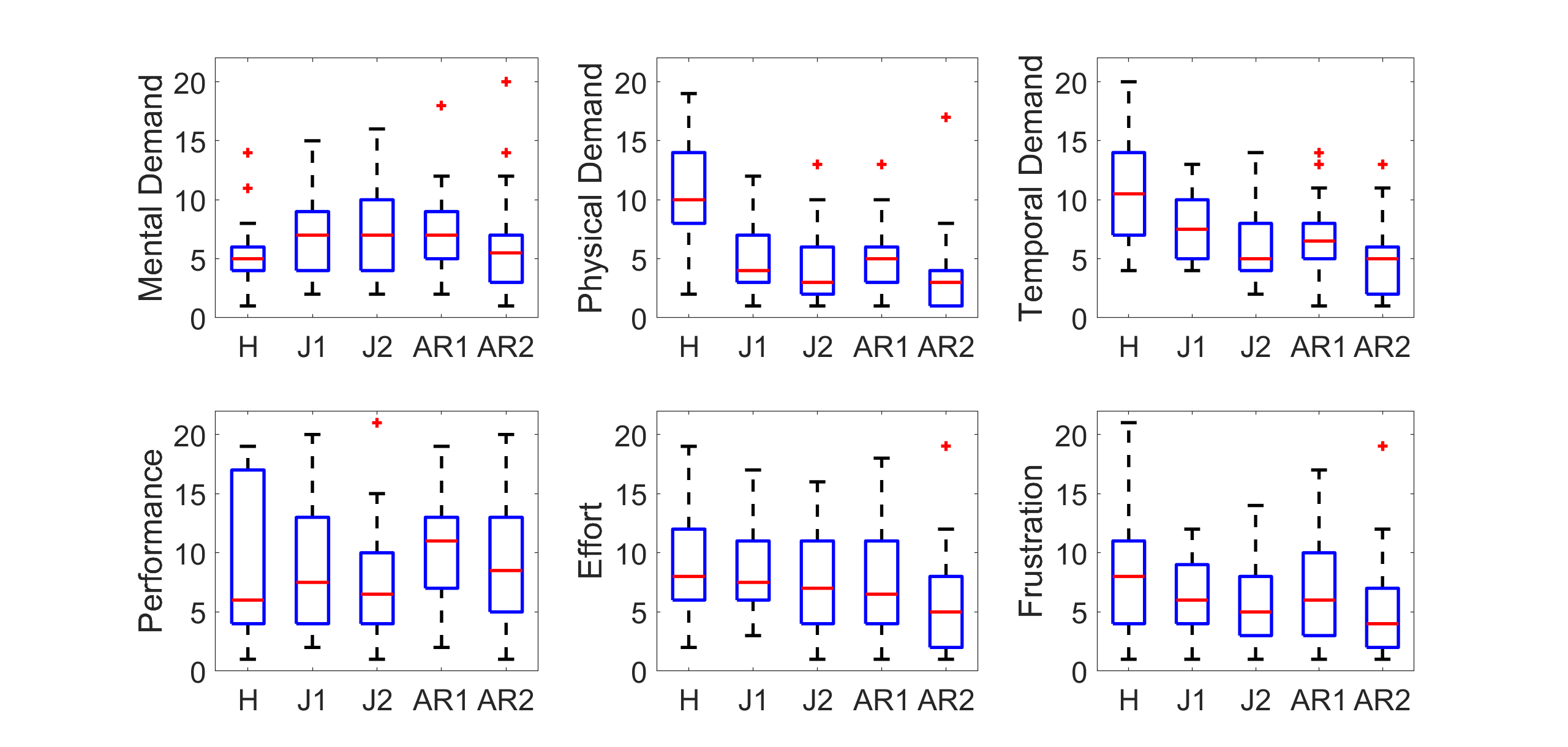}
\vspace*{-4mm}
\caption{Participant response to NASA-TLX \cite{NASA-TLX} questions (21 point scale).}
\label{fig:nasa-tlx}
\end{figure}

\textbf{Perceived Task Load.}
Table \ref{table:TLX} reports the NASA-TLX results, while Fig. \ref{fig:nasa-tlx} shows the box plots. Overall, AR2 had the smallest physical demand, temporal demand, effort and frustration scores, while AR1 had the largest performance score. H was found to have the smallest mental demand score. ANOVA indicated significant differences in physical demand ($F(4, 125)=16.74, p<0.001$), temporal demand ($F(4, 125)=10.98, p<0.001$), and effort ($F(4, 125)=2.55, p<0.05$), but not in mental demand ($F(4, 125)=1.50, p=0.206$), performance ($F(4, 125)=0.86, p=0.493$), and frustration ($F(4, 125)=1.92, p=0.112$). Post hoc analysis revealed that physical demand in J1, J2, AR1, and AR2 were significantly lower than that in H ($t(25)=6.94,p<0.001$; $t(25)=6.23,p<0.001$; $t(25)=6.50,p<0.001$; $t(25)=6.28,p<0.001$). Temporal demand in J1, J2, AR1, and AR2 were also significantly lower than H ($t(25)=4.86,p<0.001$; $t(25)=4.79,p<0.001$; $t(25)=4.71,p<0.001$; $t(25)=6.84,p<0.001$). Meanwhile, temporal demand in AR2 was significantly lower than that in AR1 ($t(25)=3.76,p<0.001$). The effort in AR2 is significantly lower than that in H ($t(25)=3.47,p<0.002$).

\begin{table}
\caption{NASA-TLX \cite{NASA-TLX} questionnaire results (21 point scale).}
\vspace*{-3mm}
\label{table:TLX}
\centering
\setlength\tabcolsep{2.5pt}
\begin{tabular}{ |c|c|c|c|c|c| } 
\hline
\textbf & \textbf{H} & \textbf{J1} & \textbf{J2} & \textbf{AR1} & \textbf{AR2} \\
\hline
% multi line cell version:
%\begin{tabular}{@{}c@{}}\textbf{mental} \\ \textbf{demand}\end{tabular}  & 5.8$\pm$2.8 & 7.1$\pm$3.2 & 7.6$\pm$4.2 & 8.2$\pm$2.8 & 6.4$\pm$3.3 \\
\textbf{mental demand} & 5.5$\pm$2.7 & 7.4$\pm$3.4 & 7.3$\pm$4.1 & 7.4$\pm$3.8 & 6.2$\pm$4.3 \\
\textbf{physical demand} & 10.3$\pm$4.5 & 5.0$\pm$2.8 & 4.2$\pm$3.1 & 5.0$\pm$2.9 & 3.5$\pm$3.3 \\  
\textbf{temporal demand} & 10.7$\pm$4.3 & 7.6$\pm$2.8 & 6.0$\pm$3.1 & 6.9$\pm$3.1 & 5.1$\pm$3.1 \\  
\textbf{performance} & 8.6$\pm$6.3 & 8.8$\pm$5.0 & 7.7$\pm$5.1 & 10.3$\pm$4.3 & 9.3$\pm$5.2 \\  
\textbf{effort} & 9.2$\pm$4.4 & 8.5$\pm$4.1 & 7.3$\pm$3.7 & 7.7$\pm$4.8 & 5.7$\pm$4.3 \\  
\textbf{frustration} & 8.3$\pm$5.3 & 6.6$\pm$3.1 & 5.5$\pm$3.5 & 6.8$\pm$4.3 & 5.5$\pm$4.5 \\  
\hline
\end{tabular}
\vspace*{-1mm}
\end{table}

\begin{figure}[]

\centering
\includegraphics[width=.45\textwidth]{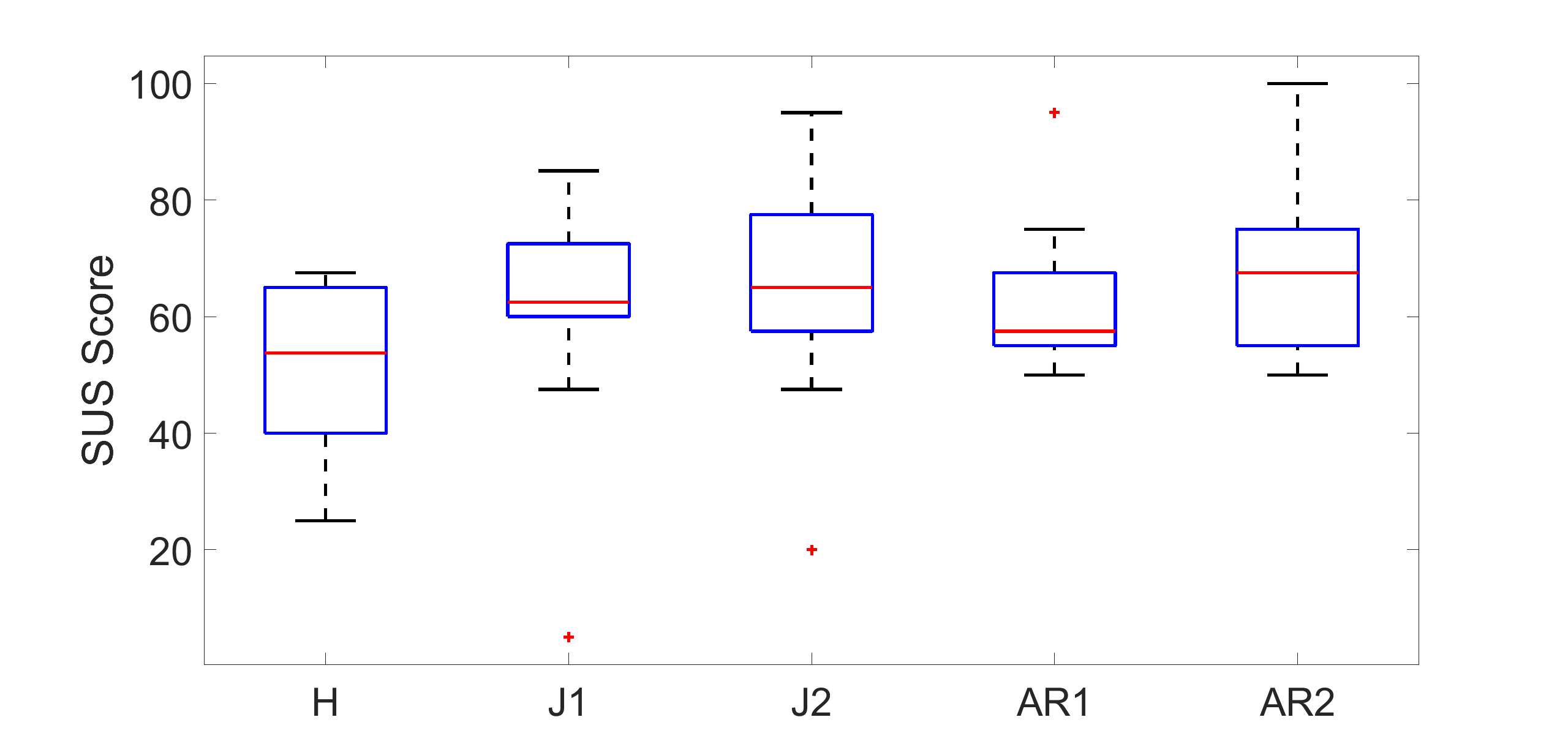}
\vspace*{-3mm}
\caption{SUS score for the five conditions}
\vspace*{-3mm}
\label{fig:sus_score}
\end{figure}

\textbf{System Usability.}
Table \ref{table:sus} shows the SUS score for the five tested conditions while Fig. \ref{fig:sus_score} show the box plot. Results showed that AR2 achieved the highest SUS score. ANOVA revealed that there were significant differences among the measured SUS score across the five conditions ($F(4,129)=3.586, p \leq 0.01$). The pairwise t-test shows a significant different between H and both J2 ($p\leq0.001$) and AR2 ($p\leq0.01$). Besides, a significant different between J1 and J2 ($p\leq0.0001$).
According to the global benchmark for SUS created by Sauro and Lewis through surveying 446 studies spanning %hardware, software, and websites,
different types of systems, the mean given score is $68 \pm 12.5$ \cite{Sauro2012}. Comparing this global mean score from the benchmark with those obtained for our five conditions tested, we found that the SUS score for H and J1 were significantly lower than the global mean ($t(25)=3.785,p=0.00086; t(25)=3.067,p=0.0051$), while the SUS score for the other conditions J2, AR1, AR2 had no significant difference from the global mean ($t(25)=0.158,p=0.876;t(25)=1.128,p=0.270;t(25)=0.441,p=0.663$). 

\begin{table}[]
\vspace*{-3mm}
\caption{System Usability Scale (SUS) scores measured in each condition.}
\vspace*{-3mm}
\label{table:sus}
\centering
\begin{tabular}{ |c|c|c|c|c|c| } 
\hline
 & \textbf{H} & \textbf{J1} & \textbf{J2} & \textbf{AR1} & \textbf{AR2} \\
\hline
\textbf{SUS Score} & $56.63\pm15.31$ & $58.08\pm16.50$ & $68.46\pm14.90$ & $64.71\pm14.87$ & $69.51\pm17.56$ \\ 
\hline
\end{tabular}
\vspace*{-1mm}
\end{table}

\begin{figure}[]

\centering
\includegraphics[width=.75\textwidth]{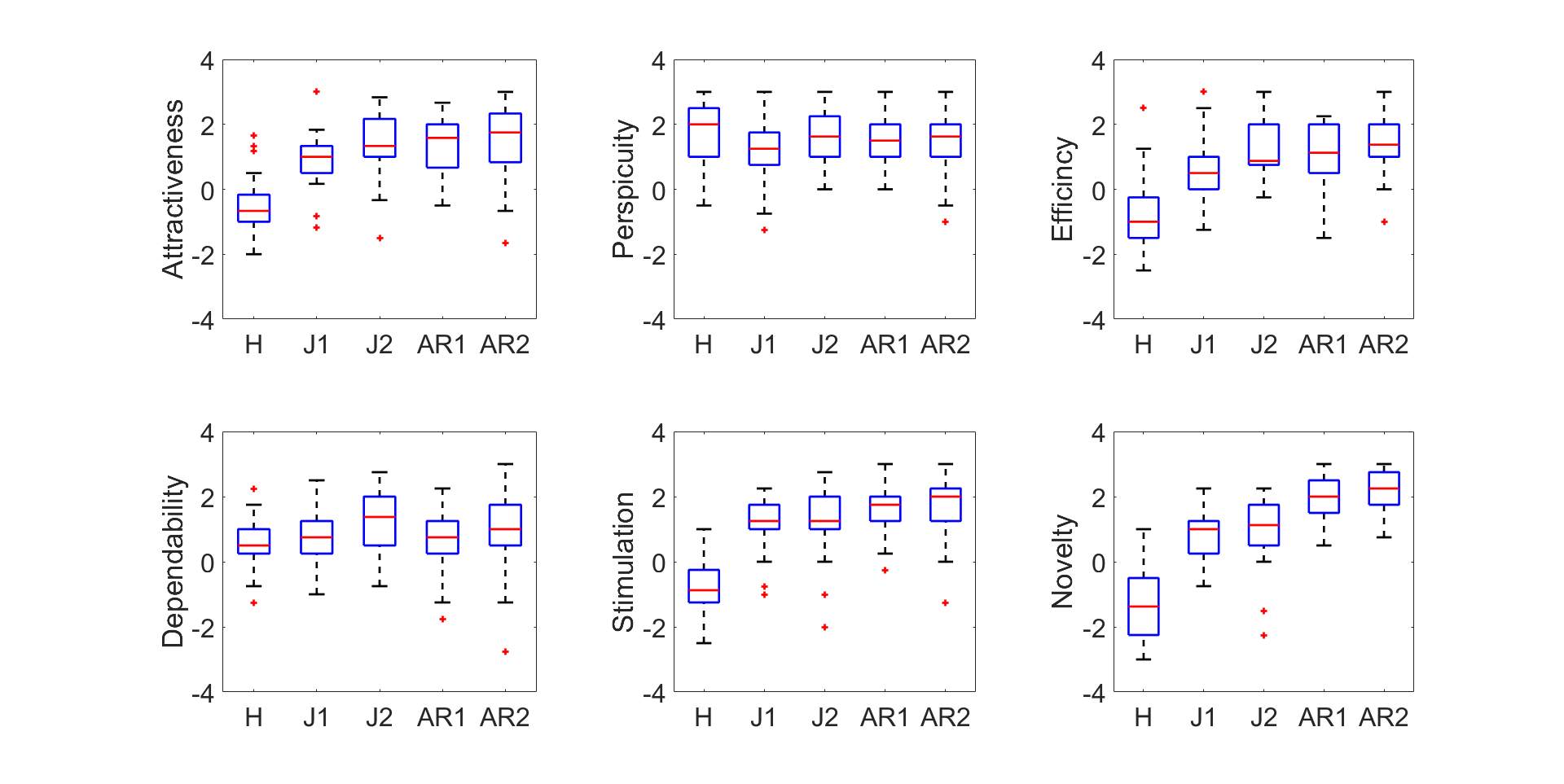}
\vspace*{-3mm}
\caption{UEQ score for the six aspects with the five conditions.}
\vspace*{-3mm}
\label{fig:ueq_score}
\end{figure}

\begin{figure}[]

\centering
\begin{subfigure}{0.85\textwidth}
    \centering
    \subcaption{H}
    \includegraphics[width=.85\textwidth]{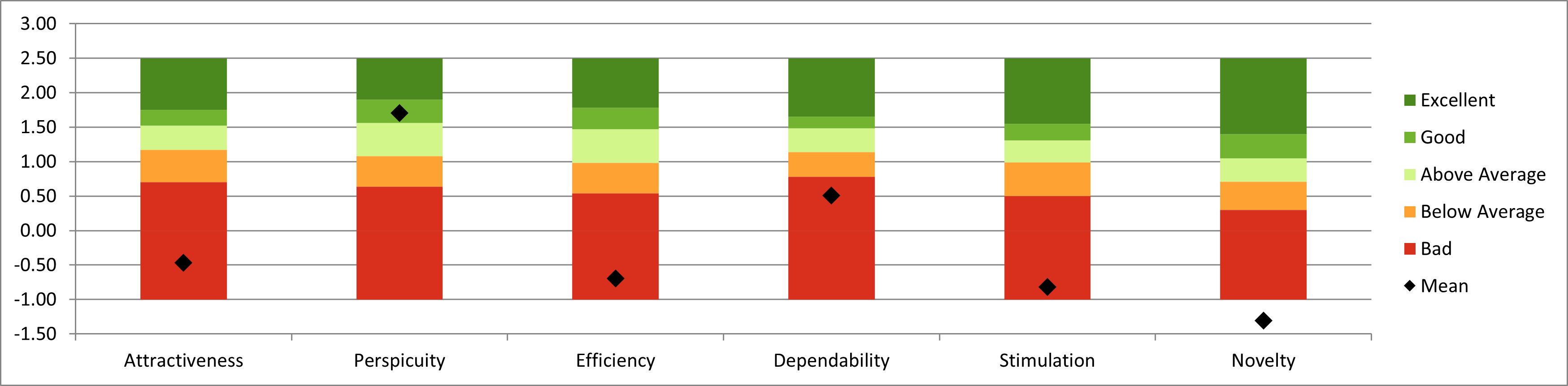}
    
\end{subfigure}
\hfill
\begin{subfigure}{0.85\textwidth}
    \centering
    \subcaption{J1}
    \includegraphics[width=.85\textwidth]{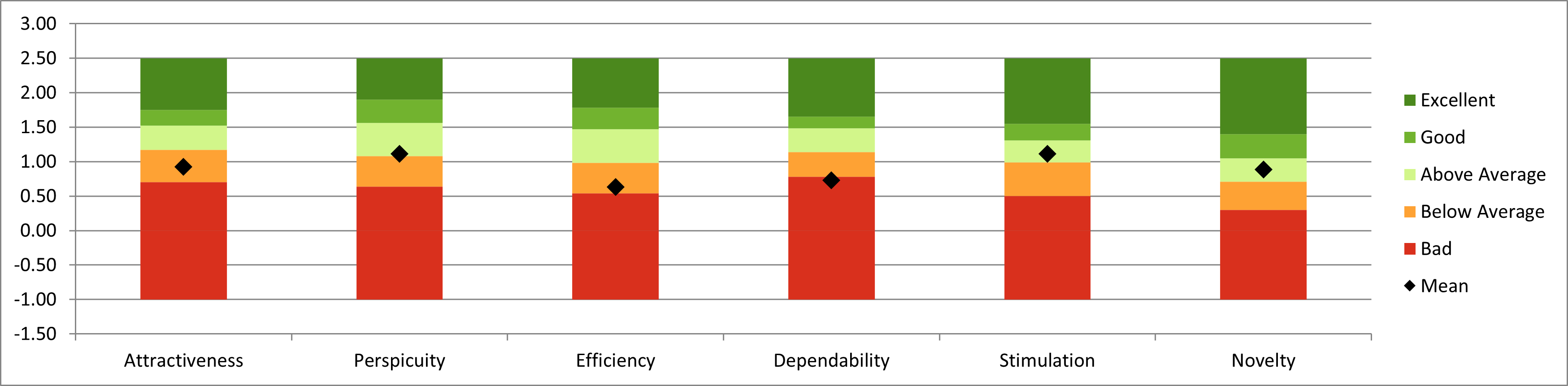}
    
\end{subfigure}
\hfill
\begin{subfigure}{0.85\textwidth}
    \centering
    \subcaption{J2}
    \includegraphics[width=.85\textwidth]{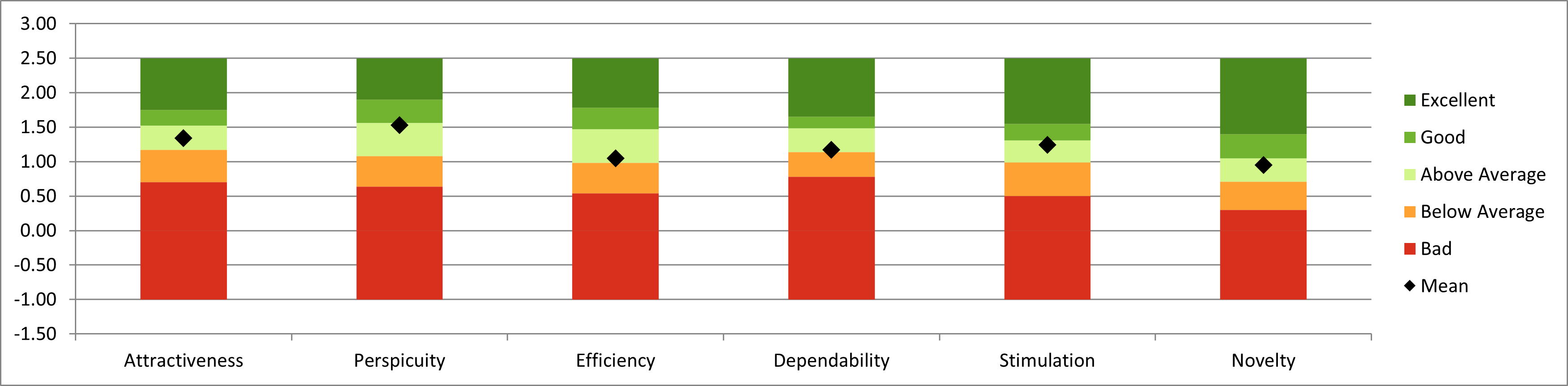}
    
\end{subfigure}
\hfill
\begin{subfigure}{0.85\textwidth}
    \centering
    \subcaption{AR1}
    \includegraphics[width=.85\textwidth]{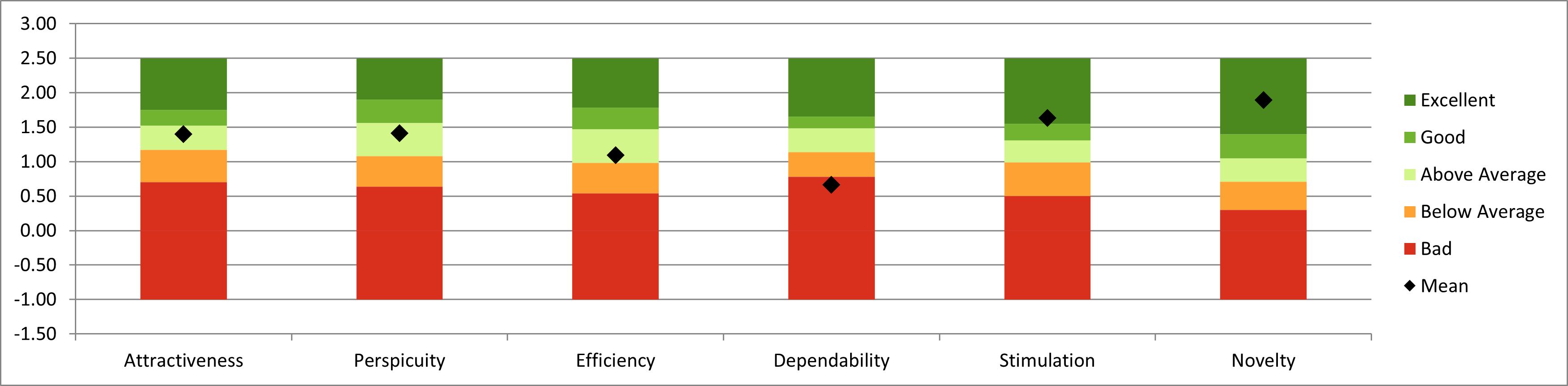}
    
\end{subfigure}
\hfill
\begin{subfigure}{0.85\textwidth}
    \centering
    \subcaption{AR2}
    \includegraphics[width=.85\textwidth]{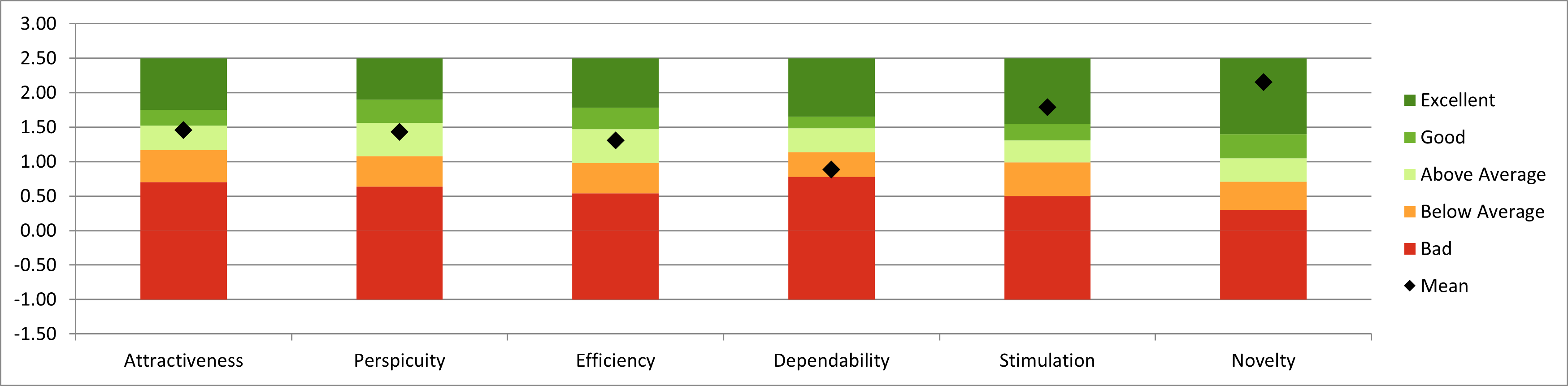}
    
\end{subfigure}
\hfill
     
\caption{UEQ scores against benchmark for the five conditions.}
\vspace*{-3mm}
\label{fig:ueq_benchmark}
\end{figure}

\textbf{User Experience.} Table \ref{table:UEQ} show the UEQ scores while Fig.\ref{fig:ueq_score} shows the box plots for the six measured aspects in the five tested conditions. Fig. \ref{fig:ueq_benchmark} shows comparisons of the UEQ scores with established benchmark \cite{Schrepp2017}. (Reported UEQ scores are after data transform \cite{UEQ}, and has a range of [-3, 3]). Overall, the AR interface achieved the highest score in Attractiveness and Novelty (AR2), and highest score in Efficiency and Stimulation (AR1). The joystick interface (J2) achieved the highest score in Dependability and Perspicuity. ANOVA indicated significant differences among the five conditions for Attractiveness ($F(4,100)=24.15 , p \leq 0.0001$), Efficiency ($F(4, 100) = 22.748 , p \leq 0.0001$), Stimulation ($ F(4, 100) = 48.057, p\leq0.0001$), and Novelty ($F(4, 100) = 73.996, p\leq0.0001$). Pairwise t-test revealed that condition H had significantly lower scores compared to the other four conditions in terms of Attractiveness ($p\leq0.0001$), Efficiency ($p\leq0.0001$), Stimulation ($p\leq0.0001$) and Novelty ($p\leq0.0001$). 
Furthermore, comparison between J1 and AR1 showed a statistical difference in terms of Stimulation ($p\leq0.05$) and Novelty ($p\leq0.001$). Similarly, comparison between J2 and AR2 showed a statistical difference in terms of Novelty ($p\leq0.001$). Also, comparison between J1 and J2 showed a statistical difference in terms of Dependability ($p\leq0.05$).
Compared against benchmarks, condition H received Bad ratings for all measures except perspicuity. Joystick conditions (J1, J2) received mostly ratings around Average (Above Average or Below Average) for the six measures. AR conditions (AR1, AR2) received Excellent ratings for stimulation and novelty, but received Below Average or Bad ratings for Dependability.

\begin{table}
\caption{UEQ scores \cite{UEQ} for each condition. Each score ranges from -3 to 3.}
\vspace*{-3mm}
\label{table:UEQ}
\centering
\setlength\tabcolsep{2.5pt}
\begin{tabular}{ |c|c|c|c|c|c| } 
\hline
\textbf & \textbf{H} & \textbf{J1} & \textbf{J2} & \textbf{AR1} & \textbf{AR2} \\
\hline
% multi line cell version:
%\begin{tabular}{@{}c@{}}\textbf{Attractiveness} \\ \textbf{Perspicuity}\end{tabular}  & 5.8$\pm$2.8 & 7.1$\pm$3.2 & 7.6$\pm$4.2 & 8.2$\pm$2.8 & 6.4$\pm$3.3 \\

\textbf{Attractiveness} & -0.47$\pm$0.91 & 0.92$\pm$0.71 & 1.34$\pm$1.01 & 1.40$\pm$0.78 & 1.46$\pm$1.50 \\
\textbf{Perspicuity} & 1.70$\pm$1.00 & 1.12$\pm$1.00 & 1.53$\pm$0.72 & 1.41$\pm$0.54 & 1.43$\pm$0.84 \\  
\textbf{Efficiency} & -0.69$\pm$1.41 & 0.64$\pm$0.83 & 1.05$\pm$0.91 & 1.10$\pm$1.06 & 1.31$\pm$0.87 \\  
\textbf{Dependability} & 0.51$\pm$0.67 & 0.73$\pm$0.82 & 1.17$\pm$1.04 & 0.66$\pm$0.99 & 0.89$\pm$1.56 \\  
\textbf{Stimulation} & -0.82$\pm$0.75 & 1.12$\pm$0.59 & 1.24$\pm$1.14 & 1.64$\pm$0.61 & 1.79$\pm$0.93 \\  
\textbf{Novelty} & -1.31$\pm$1.44 & 0.89$\pm$0.47 & 0.95$\pm$1.17 & 1.89$\pm$0.53 & 2.15$\pm$0.45 \\ 

\hline
\end{tabular}
\vspace*{-1mm}
\end{table}

% \begin{figure}[]

% \centering
% \includegraphics[width=.75\textwidth]{Images/ueq_benchmark.png}
% \vspace*{-3mm}
% \caption{UEQ scores against benchmark for the five conditions.}
% \vspace*{-3mm}
% \label{fig:ueq_benchmark}
% \end{figure}

\section{DISCUSSION}
        \label{sec:discussion}
        \subsection{Benefits of a Robotic Assistant}
In relation to our hypothesis \textbf{H1} regarding benefits of a robotic assistant, we found that regardless of user interface, in conditions J1, J2, AR1, and AR2, the introduction of a robotic assistant reduced physical and temporal demand on the participant and improved task efficiency by reducing task time, when compared to condition H. These results directly support \textbf{H1}. As current CFRP manufacturing pleating processes are still performed fully manually, these results are encouraging are they show promise that by introducing robot assistants and enabling human-robot teams in high physical-demand manufacturing tasks, we can potentially mitigate strain injuries in increase task efficiency. 

\subsection{Comparison of AR vs Joystick Interface}
In relation to our hypothesis \textbf{H2} comparing our AR interface with a standard joystick interface, we found that when using our AR interface (AR1, AR2), the completion time was significantly shorter compared to when using the joystick interface (J1, J2 respectively). These results supported the first part of \textbf{H2}, where we hypothesized that task time would be reduced. These results confirmed findings from the literature \cite{stadler,Quintero2018,ONG2020101820,Frank2016}. Inspecting the NASA-TLX results, we found that the use of our AR interface (AR2) yielded the lowest physical  and mental demand. However, neither differences in physical or mental demand reached significance compared with joystick interfaces. Hence, our results did not support the second part of \textbf{H2}, which stated that task load would be lower with the use of AR. Comparing with existing studies, some have found that use of AR reduces mental load \cite{stadler}, while others found that use of AR increases mental load and decreases physical load \cite{Quintero2018}. The discrepancy between results from our current study and existing studies may be due to different task types being considered in each study. Hence, when implementing the use of AR for robotic applications, consideration for the task at hand may be required. Existing studies tend to focus on comparing the use of AR versus alternative interfaces, but not the use of AR for different tasks. Further investigation is needed to determine how different task types influence the effects of AR on human-robot collaboration outcome.

\subsection{System Usability and User Experience}
\label{sec:discussion_sus_ueq}
Regarding hypothesis \textbf{H3} stating that AR improves system usability and user experience compared to joystick. We found that both interfaces improved SUS score when compared to human-only condition (H).
Our AR interface (AR2) achieved highest SUS score which was significantly higher than human-only condition (H). Considering the UEQ scores, the human condition received Bad ratings for all but one aspect, indicating that significant improvement is needed for current mode of manual operation. Comparing the joystick interface conditions (J1, J2), which received roughly Average ratings for all aspects, with the AR interface (AR1, AR2), we found that the AR interface has improved ratings (Excellent) for Stimulation and Novelty aspects, but worse rating (Below Average or Bad) for Dependability. Hence we found only partial support for \textbf{H3}.

It is interesting to note that the SUS scores for conditions AR2 and J2, when task division was unspecified, with participants given freedom to decide how and when to use the system, were higher than the scores for conditions AR1 and J1, when task division was predetermined, with participants told which tasks to perform and which tasks to use the robot for. This may be an indication that freedom for user to choose how to use a given system is also important towards increasing system usability, and perhaps more so than the interface itself. Conversation with our industry expert partner indeed confirms that it is important to allow the user freedom to choose how and when they like to use a robotic assistant, from a worker's perspective.

\subsection{AR Interface Promotes Robot Utilization}
Regarding hypothesis \textbf{H4} stating that robot utilization will increase with the use of AR, our results showed that with the use of our AR interface (AR2), robot utilization increased significantly. While the same is observed for J2, the increase in robot utilization in AR2 is larger than that in J2. These results provided support for \textbf{H4}. 

A more detailed examination of results revealed that our AR interface encouraged all but two participant to increase their utilization of the robot (with the remaining two participant still utilizing the robot 50\%). However, with a joystick interface, in J2, the minimum robot utilization decreased for 4 out of 26 (15\%) participants, (including 2 participants with 0\% robot utilization), despite the fact that a reduced task completion time was observed when comparing J1 to H. This meant that although utilization of robot assistant can improve task efficiency (J1 compared to H), given the joystick interface, participants may opt to not use the robot at all. This demonstrates that a beneficial system may be left abandoned by users if the interface design is inadequate. Indeed, many existing studies have pointed out the importance of designing robotics systems that are accepted and adopted by users, as it affects the continual use and further development of the technology \cite{Klamer2010, Heerink2006, Moradi2018}. Experiment results demonstrated that our AR interface can encourage the adoption and utilization of a robotic assistant technology.

\subsection{Novelty Effects}
\label{sec:discussion_novelty}
With new technologies such as AR, there are potentially novelty effects. Existing studies on the use of AR often do not examine such effects. The UEQ results from our study, however, showed that there are novelty effects form the AR interface. Participants found the AR interface to be more novel and more stimulating. They also perceived the AR interface to be less dependable, despite increased task efficiency (reduced task time) and robot utilization. These results suggested that the novelty effects may have caused a mismatch between user perception and actual task outcome. As user become more familiarized with the AR interface, we expect this mismatch to eventually fade, and task performance to further improve. 

\subsection{Participant Comments}
From the participants' feedback, we found that our AR system was generally the most preferred interface. Participants indicated that our AR interface as "easier", "faster", and "more convenient" to use, and they indicated that our AR interface was "the best one" compared to the fully manual or joystick alternatives. A few participants indicated that the use of the joystick interface was slower but provided better accuracy, as it was more difficult to position the way points  using gaze with the AR interface, and there is sometimes misalignment between the virtual and real robot. One participant had trouble with the speech commands as HoloLens' standard speech recognition was not able to accurately recognize his spoken commands. The participant expressed that this affected how much he would have used the AR system by 70\%. Although, the participant still opted to utilized the robot for more than 50\% of the task in the AR2 condition.

\subsection{Limitations}
To allow participants without expertise in CFRP fabrication to perform the experiment, we have substituted the real pleating task with a colouring task. While our experiment task simulates the key aspects of a real pleating task, comparison of our findings with literature have suggested that task type may influence human-robot interaction outcome. Hence, subsequent studies on the use of our AR system for real pleating tasks would be worthwhile. Our target task of CFRP fabrication intends to utilize large scale factory robots. While we have conducted our study using a robot test bench set up that is much larger than typical table-top robots used in existing studies \cite{stadler,Quintero2018,ONG2020101820}, the target robots at our industry partner's facility is still of much larger scale. We intend on eventually testing our system with those robots. As our study results suggested that there are novelty effects with the the use of AR, longer term studies, or studies involving longer training sessions may help better understand their effects on task performance and user experience. With increased familiarity, we expect AR to be able to further increase task efficiency and robot utilization.

In our study design, we have a partially counter-balanced ordering of the test conditions. The H condition was always first, since we wanted participants to be able to first experience the analog to the current manual method of CFRP pleating. As a result, the performance increase observed in the subsequent conditions (J1, J2, AR1, AR2), may have been due to task familiarization. In future work we would propose to ask participants to perform another H trial (H2) at the end of the experiment to measure the effect of familiarisation.
However, given the length of the experimental task it was not desirable to conduct an additional H2 trial for all participants, as we were constrained to a one hour session with participants to avoid fatigue. However, we were able to conduct an H2 trial for two of the participants to provide some insight. The resulting task completion times in H2 for these 2 participants were found to be shorter than that in H, but comparable to those in J1 and J2, and still longer than those in AR1 and AR2. This suggests that while task familiarity will improve task completion time, our AR system indeed provided benefit to these users.

\subsection{Recommendations}
\label{sec:discussion_recommendations}
Based on our study findings, we provided the following recommendations towards implementation of AR interfaces for human-robot teams.

\begin{itemize}
    \item As AR is a fairly new technology, we recommend longer training or familiarization periods to reduce novelty effects, reduce negative perception on dependability, and maximize task performance gain. 
    \item The system should permit user freedom in deciding how and when to utilized the robotic assistant, as it may otherwise negatively impact system usability.
    \item The effects of AR interfaces on task performance outcomes may depend on the actual task type. Hence consideration for the target task type should be given when considering the application of AR. For examples, in some studies, mental demand was found to increase with use of AR, while in some others it was found to decrease. Longer training and familiarization periods may potentially help reduce some of such negative effects. 

\end{itemize}
        
\section{CONCLUSION AND FUTURE WORK}
        \label{sec:conclusion}
        Toward our goal of utilizing AR for enabling intuitive collaboration in human-robot team for large-scale, labour-intensive manufacturing tasks, we have presented a study on the use of our AR interface for an experiment task simulating collaborative CFRP composite manufacturing. To our knowledge, this is the first study on the use of AR for human-robot teaming involving simultaneous collaboration in a physically shared task.
Results demonstrated that the use of AR can provide an effective user interface, increase task efficiency, reduce worker physical demand, and promote robot utilization. Compared to our previous publication \cite{chan2020AR}, this paper present a number of key new contributions, including additional experimentation, more detailed analysis, corroborating evidence, and a list of recommendations for implementing AR-based human-robot teams based on our study findings. Future studies may involving investigating how task type and novelty effects influence the performance of AR-enabled human-robot collaboration.

\begin{acks}
\label{sec:acknowledgement}
We would like to thank our collaborators Matthias Beyrle and Jan Faber at the German Aerospace Center, DLR, for their expert advice and guidance on this project. This project was supported by the Australian Research Council Discovery Projects Grant, Project ID: DP200102858.
\end{acks}

\bibliographystyle{ACM-Reference-Format}
\bibliography{ref}

\end{document}